\documentclass{article}
\usepackage{spconf,amsmath,graphicx}

\usepackage[T1]{fontenc}
\usepackage[utf8]{inputenc}
\usepackage{microtype}
\usepackage{xcolor}
\usepackage{colortbl}
\usepackage{pgfplots}
\pgfplotsset{compat=1.16}

\definecolor{perfcolor}{RGB}{200,242,255}    
\definecolor{effcolor}{RGB}{255,243,200}     
\usepackage{amsmath}
\usepackage{amssymb}
\usepackage{xspace}
\DeclareUnicodeCharacter{2212}{\ensuremath{-}}  
\usepackage{harveyballs}   

\usepackage{graphicx}
\usepackage{subcaption}

\usepackage{booktabs}
\usepackage{multirow}
\usepackage{array}
\usepackage{tabularx}
\usepackage{makecell}
\usepackage{longtable}   
\usepackage{diagbox}
\usepackage{siunitx}
\usepackage{tikz}
\usetikzlibrary{arrows.meta,positioning,shapes.geometric,shapes.misc,calc,fit,backgrounds,patterns}
\usepackage{pgfplots}
\pgfplotsset{compat=1.16}
\usepgfplotslibrary{polar,groupplots}
\usepackage{pgfplotstable}

\usepackage{algorithm}
\usepackage{algorithmic}   

\makeatletter
\newcommand{\linelabel}[1]{%
  \begingroup
    \edef\@currentlabel{\number\value{ALC@line}}
    \label{#1}%
  \endgroup
}
\makeatother

\usepackage{xcolor}
\usepackage{listings}
\definecolor{promptLightGray}{HTML}{F5F5F5}
\definecolor{promptBorderGray}{HTML}{B0B0B0}
\lstdefinestyle{promptstyle}{
  basicstyle=\ttfamily\scriptsize,
  breaklines=true,
  breakatwhitespace=false,
  backgroundcolor=\color{promptLightGray},
  frame=single,
  framerule=0.8pt,
  rulecolor=\color{promptBorderGray},
  showspaces=false,
  showstringspaces=false,
  showtabs=false,
  xleftmargin=0.8pt,
  xrightmargin=0.8pt,
  resetmargins=true,
  breakautoindent=false,
  columns=fullflexible,
  keepspaces=true,
  numbers=none
}

\usepackage{enumitem}
\usepackage{comment}
\usepackage{verbatim}
\usepackage{fancyvrb}
\usepackage{adjustbox}
\usepackage{pifont}
\newcommand{\cmark}{\ding{51}}
\newcommand{\xmark}{\ding{55}}

\usepackage[hidelinks]{hyperref}
\usepackage{xurl}

\newcommand{\method}{\textsc{PatchBandit}\xspace}

\title{Budget-Aware Adaptive Adversarial Patches for Black-Box Object Detection}
\name{Pedram MohajerAnsari, Amir Salarpour, David Fernandez, Mert D. Pesé}
\address{School of Computing, Clemson University, USA\\
         \{pmohaje, asalarp, dferna3, mpese\}@clemson.edu}
\begin{document}
%
\maketitle

\begin{abstract}
Adversarial patches pose a practical threat to modern object detectors. Prior work shows vulnerability, but three gaps limit actionable insight: (i) few \emph{score-based black-box} attacks \emph{jointly} optimize patch \emph{location, texture, and size} under tight query budgets; (ii) success is rarely tied to the patch's \emph{visual footprint}; and (iii) evaluations often conflate EOT robustness with plain-view suppression. We present \method{}, a query-efficient, budget-adaptive black-box attack that couples a lightweight \emph{Contextual Thompson-Sampling} placer with NES-style pixel updates, growing the patch only when progress stalls. Reporting is anchored by a \emph{strict plain-image} suppression test; EOT is audited but never used as a substitute for success, and optional appearance/printability weights expose strength--visibility trade-offs. Across YOLOv5, Faster R-CNN, and YOLOS, \method{} achieves strong suppression on CNN-based detectors and substantial suppression on the transformer-based detector, using compact patches and exposing clear query--footprint trade-offs relative to fixed-size and heuristic baselines. A print--capture pilot further shows transfer across unseen physical objects and viewpoints. Code is available at \textcolor{blue}{\url{https://github.com/pedram-mohajer/PatchBandit}}.
\end{abstract}


\section{Introduction}
\label{sec:intro}

In critical applications such as autonomous driving, robotics, and surveillance, computer vision systems increasingly depend on object detectors to detect and classify objects within a scene. These detectors span one-stage CNN architectures such as YOLO~\cite{redmon2016you}, two-stage CNN detectors such as Faster R-CNN~\cite{ren2015faster}, and transformer-based detectors such as YOLOS~\cite{fang2021you}. Like image classifiers, these detectors are vulnerable to adversarial perturbations---small, crafted input changes that induce incorrect predictions~\cite{szegedy2013intriguing}. Among the most practical threats are adversarial patches~\cite{brown2017adversarial}: localized patterns that can be printed and placed in the environment so cameras repeatedly capture them and downstream models are consistently misled. Because a single patch can be reused, scaled, and repositioned across scenes, patch attacks align naturally with physical-world threat scenarios.

\begin{figure}[t]
  \centering
  \setlength{\abovecaptionskip}{4pt}
  \setlength{\belowcaptionskip}{4pt}
  \captionsetup[subfigure]{aboveskip=2pt, belowskip=2pt}
  \begin{subfigure}[t]{0.35\linewidth}
    \centering
    \includegraphics[width=\linewidth,height=4cm,keepaspectratio]{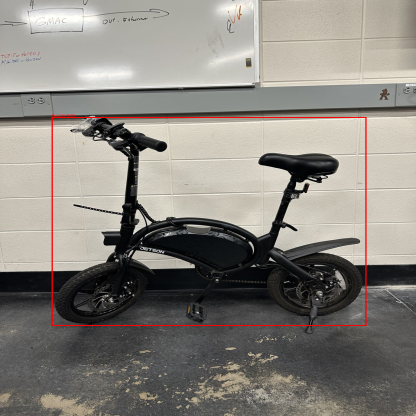}
    \subcaption{Clean physical}
    \label{fig:qual:phys_clean}
  \end{subfigure}\hspace{0.5em}
  \begin{subfigure}[t]{0.35\linewidth}
    \centering
    \includegraphics[width=\linewidth,height=4cm,keepaspectratio]{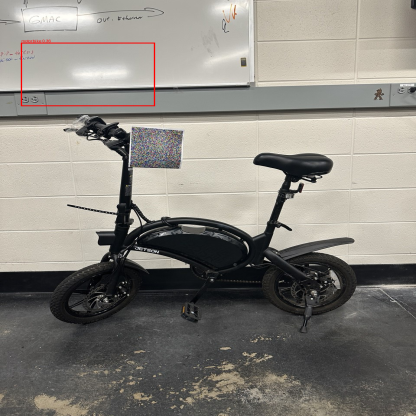}
    \subcaption{Patched physical}
    \label{fig:qual:phys_patched}
  \end{subfigure}%
  
  \begin{subfigure}[t]{0.35\linewidth}
    \centering
    \includegraphics[width=\linewidth,height=4cm,keepaspectratio]{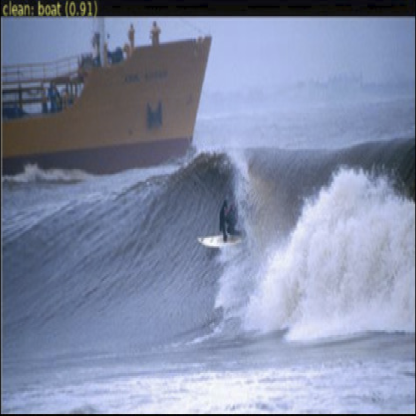}
    \subcaption{Clean digital}
    \label{fig:qual:dig_clean}
  \end{subfigure}\hspace{0.5em}
  \begin{subfigure}[t]{0.35\linewidth}
    \centering
    \includegraphics[width=\linewidth,height=4cm,keepaspectratio]{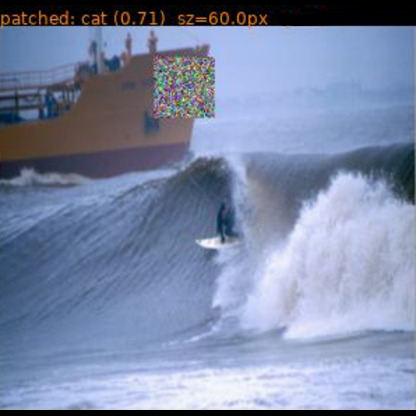}
    \subcaption{Patched digital}
    \label{fig:qual:dig_patched}
  \end{subfigure}
  \caption{A 160 px physical patch flips the top YOLO prediction from \textit{bicycle} (0.95) to \textit{motorbike} (0.36); a 60 px digital patch flips a clean \textit{boat} (0.91) to \textit{cat} (0.92).}
  \label{fig:qual_single}
\end{figure}

However, existing patch methods leave a gap between what is demonstrated in controlled settings and what is feasible against deployed detectors. Prior work includes universal white-box patches~\cite{brown2017adversarial,liu2018dpatch}, physical demonstrations~\cite{thys2019fooling}, and query-efficient black-box attacks on classifiers~\cite{ilyas2018black,ilyas2019bandit}, yet many real detectors are score-only black boxes~\cite{liang2022parallel}, which limits access to gradients and internal signals. This challenge applies across detector architectures, from CNN-based detectors to transformer-based models such as YOLOS.

Under this interface, three issues remain. First, attackers must jointly decide where to place a patch and what texture to optimize under a tight query budget, but many approaches fix location, assume loose budgets, or rely on surrogate gradients~\cite{wei2022simopt}. Second, success is rarely reported as a function of patch area or appearance cost, despite the practical importance of visual footprint~\cite{lapid2023patchinvisibility}. Third, Expectation-over-Transformation (EOT) training is common~\cite{athalye2018synthesizing}, but EOT-averaged progress can diverge from suppression on the plain camera view encountered in deployment~\cite{yang2020patchattack}, motivating evaluation that separates plain-view success from robustness across transforms.


\begin{table*}[t]
  \centering
  \tiny
  \setlength{\tabcolsep}{1.5pt}
  \renewcommand{\arraystretch}{0.9}
\caption{\textbf{Representative adversarial patch attacks.}
Published results are reported for prior methods, while \method{} is evaluated under our protocol.
WB = white-box; BB = score-only black-box; cls = image classification; det = object detection.}
  \label{tab:patch_compare}

  \resizebox{\textwidth}{!}{%
    \begin{tabular}{@{} l c c c c p{0.23\textwidth} p{0.23\textwidth} @{}}
      \toprule
      \textbf{Method} 
        & \textbf{Threat} 
        & \multicolumn{1}{c}{\textbf{Phys.}} 
        & \textbf{Task} 
        & \textbf{Atk} 
        & \textbf{Reported effect / protocol}
        & \textbf{Notes} \\
      \midrule

      Brown17~\cite{brown2017adversarial}
        & WB & \harveyBallFull & cls & T
        & 98--99\% top-1 targeted success (digital); 93\% at 1\,m (print)
        & Single reusable patch trained with EOT against Inception \& VGG. \\[2pt]
      \midrule

      DPATCH18~\cite{liu2018dpatch}
        & WB & \harveyBallNone & det & B
        & VOC07 mAP $\downarrow$ 65.70\% $\to$ 0.00\% (U); patch-only detections (T)
        & $40\times40$\,px universal patch; cross-model transfer; digital-only. \\[2pt]
      \midrule

      Thys19~\cite{thys2019fooling}
        & WB & \harveyBallFull & det & U
        & YOLOv2 person AP $\downarrow$ 100\% $\to$ 25.5\%
        & $40\times40$\,cm printed patch; needs torso placement. \\[2pt]
      \midrule

      Zolfi21~\cite{zolfi2021translucent}
        & WB & \harveyBallFull & det & U
        & Stop-sign miss rate 42.3\% (physical); AP $\downarrow$ 95.17\% $\to$ 52.71\%
        & Translucent lens sticker; hides a selected class without scene changes. \\[2pt]
      \midrule

      Lapid23~\cite{lapid2023patchinvisibility}
        & BB & \harveyBallNone & det & U
        & AP 23.2\% ($\downarrow$ 43 pp) within 110\,k queries
        & GAN-style universal patch; query-efficient NES; no physical test yet. \\[2pt]
      \midrule

      PATCHBANDIT (ours)
        & BB & \harveyBallFull & det & U
        & 77.5\% strict suppression under our protocol; 8.3\% median patch area
        & Instance-specific, size-adaptive patch; bandit placer + NES. \\
      \bottomrule

      \multicolumn{7}{l}{%
        \harveyBallFull\ Yes \quad
        \harveyBallNone\ No \quad
        \hfill
        Atk:\; T = Targeted,\; U = Untargeted,\; B = Both
      }
    \end{tabular}%
  }
\end{table*}

We present \method{}, a score-based black-box adversarial patch attack designed around these constraints. \method{} interleaves a lightweight contextual Thompson-sampling placer to efficiently explore placement cells with NES-style zeroth-order texture updates, and grows patch size only when progress stalls, exposing explicit success--versus--area trade-offs. Evaluation is anchored by a strict plain-image suppression criterion; EOT is audited and logged, but never substituted for success. We evaluate \method{} across three detector families: YOLOv5 as a one-stage CNN detector, Faster R-CNN as a two-stage CNN detector, and YOLOS as a transformer-based detector. On YOLOv5, \method{} achieves 77.5\% strict suppression with compact patches (Figure~\ref{fig:qual_single}); on Faster R-CNN, it reaches 89.7\% strict suppression with very small median patch area; and on YOLOS, it achieves 59.1\% strict suppression, indicating that the attack remains effective but is less pronounced on the transformer-based detector. A print--capture pilot further indicates that digitally optimized patches can transfer across viewpoints and object instances, repeatedly inducing mis-detections.

This paper makes the following contributions:
\begin{itemize}[leftmargin=*,itemsep=1pt,topsep=2pt]
  \item A score-based black-box detector patch attack that jointly optimizes placement, texture, and size under explicit query budgets by combining contextual Thompson sampling, NES updates, and a progress-triggered growth policy.
  \item A footprint-aware reporting protocol that prioritizes strict plain-image suppression and reports EOT robustness separately, enabling clear success--visibility trade-off analysis (including \(J=q_{\text{med}}\!\times\!a_{\text{med}}\)).
  \item An empirical evaluation on YOLOv5, Faster R-CNN, and YOLOS showing strong suppression on CNN-based detectors and substantial suppression on a transformer-based detector, plus a print--capture pilot demonstrating transfer across physical viewpoints and objects.
\end{itemize}

\section{Related Work}
\label{sec:related}

\noindent\textbf{Adversarial Patches and Physical Attacks on Vision Models.}
Adversarial patches are localized visual triggers that induce model misbehavior when inserted into a scene. Brown \textit{et al.}\ introduced \emph{universal, printable} patches that drive classifiers toward a chosen target under varied viewing conditions~\cite{brown2017adversarial}. Liu \textit{et al.}\ extended this idea to object detectors with \textsc{DPatch}, degrading both classification and localization in Faster R-CNN and YOLO~\cite{liu2018dpatch}. Thys \textit{et al.}\ demonstrated wearable/held printed patches that reduce person-detection confidence in real-world surveillance settings~\cite{thys2019fooling}. Zolfi \textit{et al.}\ proposed a \emph{translucent} lens-mounted sticker that suppresses stop-sign detections without modifying the sign itself~\cite{zolfi2021translucent}.

\noindent\textbf{Black-Box and Query-Efficient Adversarial Example Generation.}
Early score-based black-box attacks estimated gradients via \emph{zeroth-order} optimization (e.g., ZOO), avoiding substitute models~\cite{chen2017zoo}. Ilyas \textit{et al.}\ improved query efficiency with Natural Evolution Strategies (NES)~\cite{ilyas2018blackbox} and later ``Bandits \& Priors,'' which leverages structured gradient priors through bandit optimization to further reduce queries~\cite{ilyas2019bandit}. Randomized, region-wise methods such as the Square Attack provided another query-efficient alternative for high-dimensional inputs by using structured, localized updates~\cite{andriushchenko2020square}.

\noindent\textbf{Joint Optimization of Patch Content and Placement.}
Patch effectiveness depends on both \emph{what} is printed and \emph{where} it is placed. Prior work has treated placement as a decision variable, using RL or black-box optimization to jointly adapt position and appearance under query budgets~\cite{wei2022simopt}. Other approaches use evolutionary/annealing-style searches in camouflage-style pipelines (CamoPatch)~\cite{huang2020universal}, and recent score-based detector attacks optimize patches at locations via zeroth-order updates (BB-Patch)~\cite{kumar2024bbpatch}. Earlier detector work such as \textsc{DPatch} observed some location invariance but did not perform dense, adaptive placement search~\cite{liu2018dpatch}. Building on this line, our contextual bandit provides a lightweight, query-aware policy for selecting promising placement cells without full RL training.

\noindent\textbf{Stealth, Camouflage, and Printability Constraints.}
Because effective patches are often visually conspicuous, prior work constrains appearance for camouflage and physical realizability. Camo-style evolutionary methods encourage texture-matching while remaining adversarial~\cite{huang2020universal}, and Patch of Invisibility uses generative priors to produce natural-looking, printable black-box detector patches~\cite{lapid2023patchinvisibility}. Wearable person-detector attacks similarly highlight the trade-off between attack strength (often high-contrast) and visual plausibility~\cite{thys2019fooling}.

\noindent\textbf{Evaluation Protocols, EOT, and Physical Robustness.}
A key challenge is ensuring patches remain effective under real-world transformations (viewpoint, scale, lighting). Athalye \textit{et al.}\ introduced \emph{Expectation Over Transformation} (EOT) to optimize perturbations over a transformation distribution~\cite{athalye2018synthesizing}. Eykholt \textit{et al.}\ showed that digital gains can collapse in physical settings without rigorous robustness-oriented evaluation~\cite{eykholt2018robustphysical}. Detector patch work such as \textsc{DPatch} and Patch of Invisibility similarly tests under diverse, realistic environmental variation, though heavy augmentation can obscure whether the plain image is actually suppressed~\cite{liu2018dpatch,lapid2023patchinvisibility}.

\noindent\textbf{Positioning of \method{}.} 
We target \emph{score-based black-box} detectors and combine three elements that earlier studies explored in isolation: a lightweight contextual bandit selects promising cells under a query budget, NES performs gradient-free pixel updates, and a budget-adaptive size ladder enlarges the patch only when progress stalls. Strict success is recorded on the plain image, while EOT behavior is logged separately; the attack loop records queries, area, and confidence trajectories for reproducible ablations.

Table~\ref{tab:patch_compare} places this design beside prior work. \cite{brown2017adversarial}, \cite{liu2018dpatch}, \cite{thys2019fooling} and \cite{zolfi2021translucent} rely on white-box gradients to train large, universal overlays that require fixed footprints and long optimization.  \cite{lapid2023patchinvisibility} move to the same score-only interface but use a GAN latent search that lowers Tiny-YOLOv3 average precision to 23 \% at a cost of $\sim$~1.1×10\textsuperscript{5} queries.  In the same setting, \method{} suppresses the original top class on 77.5 \% of images with a median of 49 calls and an 8.3 \% patch area—over three orders of magnitude fewer queries and a smaller footprint.  White-box universals remain useful when gradients are revealed or a single reusable sticker is desired, yet that threat model is less representative of commercial detectors.  Under the more restrictive but realistic score-only interface, our results show that near-complete suppression is attainable with limited queries and an instance-specific, size-adaptive patch. Experiments with the patch placed on a bottle confirm that the attack transfers across varying viewing distances and angles.

\section{Methodology}
\label{sec:method}

Algorithm~\ref{alg:method} summarizes \method, a query-bounded black-box patch attack that jointly optimizes (i) where to place a patch, (ii) the patch texture, and (iii) the patch size. Each iteration selects a candidate location via contextual Thompson sampling, updates patch pixels with NES-style zeroth-order optimization, and grows the patch when progress stalls.

\noindent\textbf{Threat model and objective.}
We consider a score-based black-box detector $f$ queried on an RGB image $I\in[0,1]^{H\times W\times 3}$, returning detections $\{(b_j,c_j,s_j)\}$ with confidences $s_j\in[0,1]$. For any class $c$, let $c(I,c)=\max_{j:c_j=c}s_j$ denote its maximum confidence (0 if absent). We define the initial top class as $t=\arg\max_c c(I,c)$ with clean score $s_0=c(I,t)$. The attacker chooses a square patch $\theta\in[0,1]^{3\times m\times m}$ and a paste location $\ell=(x,y)$, obtaining $I'=\textsc{Paste}(I,\theta,\ell)$. Success is measured by strict suppression of the original top class, using only score queries to $f$ (no gradients or parameters).

\noindent\textbf{Patch pasting and constraints.}
Given $\theta$ and top-left $\ell$, we overwrite the corresponding image crop; candidates that spill outside the image are discarded. In digital mode we clamp $\theta\in[0,1]$; in printable mode we clamp to $[\theta_{\min},\theta_{\max}]$.

\noindent\textbf{Candidate locations and context features.}
We discretize the image into a $G\times G$ grid of feasible top-left locations $\mathcal{L}$. Each location $\ell$ has a lightweight context vector $\phi(\ell)\in\mathbb{R}^4$ capturing proximity/overlap with the nearest ground-truth box, local texture variance, and whether the cell touches the border. These features bias search toward object-bearing, high-contrast regions that tend to yield stronger suppression~\cite{liu2018dpatch}.

\noindent\textbf{Contextual Thompson sampling for placement.}
We model the per-location reward as a linear function of context,
$r = \langle w_\star, \phi(\ell)\rangle + \xi$ with $\xi\sim\mathcal{N}(0,\sigma^2)$ and prior $w\sim\mathcal{N}(0,\lambda^{-1}I)$. At iteration $t$, we sample $\tilde w\sim\mathcal{N}(\mu_t,\Sigma_t)$ from the posterior and select
$\ell_t=\arg\max_{\ell\in\mathcal{L}}\langle \tilde w, \phi(\ell)\rangle$, then update the posterior with the observed reward. (We use $w$ for bandit weights to avoid overloading the patch symbol $\theta$.)

\noindent\textbf{NES-style patch update.}
With the chosen location fixed, we update patch pixels via Natural Evolution Strategies (NES)~\cite{ilyas2018blackbox}. Each iteration samples $M$ perturbations $\epsilon_i\sim\mathcal{N}(0,I)$, evaluates the reward for candidate patches $\theta^{(i)}=\Pi(\theta+\sigma_{\text{nes}}\epsilon_i)$, and applies a stochastic ascent step to improve $\theta$ under the same clamping operator $\Pi(\cdot)$.

\noindent\textbf{Reward and appearance terms.}
For the core digital setting shown in Alg.~\ref{alg:method}, the reward is the reduction in the original top-class confidence:
$r(\theta,\ell)= s_0 - c(\textsc{Paste}(I,\theta,\ell),t)$.
When desired, we augment this objective with (i) an EOT-averaged confidence term~\cite{athalye2018synthesizing} and (ii) optional appearance/printability penalties to expose strength--visibility trade-offs.

\noindent\textbf{Adaptive size ladder.}
We use a size ladder $\mathcal{S}=\{m_1<\cdots<m_L\}$. Optimization starts at $m_1$ and tracks the best reward so far; if reward fails to improve for $\tau_{\text{grow}}$ iterations, we upsample $\theta$ to the next size and rebuild the placement grid. This grows patch area only when needed.

\noindent\textbf{Query budget and strict success.}
Each iteration evaluates $M$ patch candidates (and optionally additional transformed views); the run halts when the query cap is reached. We declare success only when the learned patch reduces the original top-class confidence by at least $\Delta_{\min}$ and pushes it below an absolute cap $\tau_{\text{abs}}$:
\begin{equation}
(s_0 - c_{\text{plain}}(\theta,\ell))\ge\Delta_{\min}
\quad\land\quad
c_{\text{plain}}(\theta,\ell)\le\tau_{\text{abs}}.
\label{eq:strictsuccess}
\end{equation}
We use $\Delta_{\min}=0.25$ and $\tau_{\text{abs}}=0.10$ in all experiments.

\begin{algorithm}[t]
\caption{\method: Contextual Bandit + NES Patch Attack}
\label{alg:method}
\footnotesize
\begin{algorithmic}[1]
\REQUIRE Image $I$, detector $f$, iterations $T$, size ladder $\mathcal{S}=\{m_1<\cdots<m_L\}$, growth patience $\tau_{\text{grow}}$, NES $(M, \sigma_{\text{nes}}, \eta)$, success thresholds $(\Delta_{\min}, \tau_{\text{abs}})$
\STATE Target class $t \gets \arg\max f(I)$ with score $s_0$; init $\theta \sim \mathcal{U}[0,1]^{3\times m_1\times m_1}$, grid $\mathcal{L}$ with features $\phi(\ell)\in\mathbb{R}^4$, bandit posterior $(A,b)$, stall counter
\FOR{$\text{iter}=1$ \TO $T$}
    \STATE Sample $\tilde{w}\sim\mathcal{N}(A^{-1}b,\,\sigma^2 A^{-1})$; pick $\ell\gets\arg\max_{\ell'\in\mathcal{L}}\langle\tilde{w},\phi(\ell')\rangle$
    \STATE Draw $\{\epsilon_i\}_{i=1}^{M}\sim\mathcal{N}(0,I)$; evaluate rewards $r_i \gets s_0-c\bigl(f(\textsc{Paste}(I,\Pi(\theta+\sigma_{\text{nes}}\epsilon_i),\ell)),t\bigr)$
    \STATE Update $\theta\gets\Pi(\theta+\tfrac{\eta}{M}\sum_i \tilde{r}_i\epsilon_i)$ \COMMENT{$\tilde{r}_i$ z-scored}
    \STATE $r^\star\gets\max_i r_i$; if $(s_0-c^\star)\ge\Delta_{\min}$ and $c^\star\le\tau_{\text{abs}}$: \RETURN $(\theta,\ell)$
    \STATE If no improvement for $\tau_{\text{grow}}$ iters and $m<m_L$: grow patch via \textsc{Upsample} and rebuild $\mathcal{L}$
    \STATE Update bandit: $A\gets A+\phi(\ell)\phi(\ell)^\top$, $b\gets b+r^\star\phi(\ell)$
\ENDFOR
\RETURN best $(\theta,\ell)$ found
\end{algorithmic}
\end{algorithm}

\section{Evaluation}
\label{sec:experiment}

\subsection{Digital Results}

\noindent\textbf{Experimental palette.}
We evaluate 18 controlled variants of \method{} that each toggle a single design knob (growth patience, query schedule, context masking, budget stress tests, stealth bias, EOT robustness, and fixed-size controls). Unless stated otherwise, results use the same protocol as the baseline; selected settings are repeated across seeds to assess variability.

{\setlength{\tabcolsep}{2pt}%
\begin{table*}[t]
  \centering
  \scriptsize
   \caption{Digital attack performance and efficiency. \textbf{Performance columns} (blue): Strict success (\%), confidence drop, patch area (\%), queries. \textbf{Efficiency columns} (orange): queries/success, area/success, joint cost $J_{\text{frac}} = q_{\text{med}} \times (a_{\text{med}}/100)$.}
  \label{tab:combined}
  \resizebox{\textwidth}{!}{%
  \begin{tabular}{@{}l|cccc|ccc|cccc|ccc|cccc|ccc@{}}
    \toprule
    & \multicolumn{7}{c|}{\textbf{YOLO} (one-stage CNN)}
    & \multicolumn{7}{c|}{\textbf{Faster R-CNN} (two-stage CNN)}
    & \multicolumn{7}{c}{\textbf{YOLOS} (ViT transformer)} \\
    \cmidrule(lr){2-8}\cmidrule(lr){9-15}\cmidrule(lr){16-22}
    \multirow{2}{*}{\textbf{Variant}} 
      & \multicolumn{4}{c|}{\cellcolor{perfcolor!50}\textbf{Performance}} 
      & \multicolumn{3}{c|}{\cellcolor{effcolor!50}\textbf{Efficiency}}
      & \multicolumn{4}{c|}{\cellcolor{perfcolor!50}\textbf{Performance}} 
      & \multicolumn{3}{c|}{\cellcolor{effcolor!50}\textbf{Efficiency}}
      & \multicolumn{4}{c|}{\cellcolor{perfcolor!50}\textbf{Performance}}
      & \multicolumn{3}{c}{\cellcolor{effcolor!50}\textbf{Efficiency}} \\
    \cmidrule(lr){2-5}\cmidrule(lr){6-8}\cmidrule(lr){9-12}\cmidrule(lr){13-15}\cmidrule(lr){16-19}\cmidrule(lr){20-22}
    & \cellcolor{perfcolor!30}Strict & \cellcolor{perfcolor!30}Drop$_t$ & \cellcolor{perfcolor!30}Area & \cellcolor{perfcolor!30}Q.
    & \cellcolor{effcolor!30}Q./s & \cellcolor{effcolor!30}A./s & \cellcolor{effcolor!30}$J_{\text{frac}}$
    & \cellcolor{perfcolor!30}Strict & \cellcolor{perfcolor!30}Drop$_t$ & \cellcolor{perfcolor!30}Area & \cellcolor{perfcolor!30}Q.
    & \cellcolor{effcolor!30}Q./s & \cellcolor{effcolor!30}A./s & \cellcolor{effcolor!30}$J_{\text{frac}}$
    & \cellcolor{perfcolor!30}Strict & \cellcolor{perfcolor!30}Drop$_t$ & \cellcolor{perfcolor!30}Area & \cellcolor{perfcolor!30}Q.
    & \cellcolor{effcolor!30}Q./s & \cellcolor{effcolor!30}A./s & \cellcolor{effcolor!30}$J_{\text{frac}}$ \\
    \midrule
    \method{} (ctx-strict)
      & \cellcolor{perfcolor!10}77.5 & \cellcolor{perfcolor!10}0.452 & \cellcolor{perfcolor!10}8.32 & \cellcolor{perfcolor!10}1079
      & \cellcolor{effcolor!10}63.26 & \cellcolor{effcolor!10}10.74 & \cellcolor{effcolor!10}6.794
      & \cellcolor{perfcolor!10}89.7 & \cellcolor{perfcolor!10}0.075 & \cellcolor{perfcolor!10}0.92 & \cellcolor{perfcolor!10}177
      & \cellcolor{effcolor!10}8.92 & \cellcolor{effcolor!10}1.03 & \cellcolor{effcolor!10}\textbf{0.092}
      & \cellcolor{perfcolor!10}59.1 & \cellcolor{perfcolor!10}0.467 & \cellcolor{perfcolor!10}8.32 & \cellcolor{perfcolor!10}290
      & \cellcolor{effcolor!10}149.95 & \cellcolor{effcolor!10}4.58 & \cellcolor{effcolor!10}24.089 \\
    
    GP5 (fast growth)
      & \cellcolor{perfcolor!10}77.7 & \cellcolor{perfcolor!10}0.455 & \cellcolor{perfcolor!10}8.32 & \cellcolor{perfcolor!10}529
      & \cellcolor{effcolor!10}30.88 & \cellcolor{effcolor!10}10.71 & \cellcolor{effcolor!10}3.307
      & \cellcolor{perfcolor!10}90.0 & \cellcolor{perfcolor!10}0.080 & \cellcolor{perfcolor!10}2.08 & \cellcolor{perfcolor!10}155
      & \cellcolor{effcolor!10}7.78 & \cellcolor{effcolor!10}2.31 & \cellcolor{effcolor!10}0.180
      & \cellcolor{perfcolor!10}60.8 & \cellcolor{perfcolor!10}0.558 & \cellcolor{perfcolor!10}14.79 & \cellcolor{perfcolor!10}47.5
      & \cellcolor{effcolor!10}32.88 & \cellcolor{effcolor!10}8.13 & \cellcolor{effcolor!10}7.027 \\
    
    Nocx\_Strict $\varepsilon$-greedy
      & \cellcolor{perfcolor!10}78.9 & \cellcolor{perfcolor!10}0.406 & \cellcolor{perfcolor!10}8.32 & \cellcolor{perfcolor!10}1013
      & \cellcolor{effcolor!10}58.32 & \cellcolor{effcolor!10}10.55 & \cellcolor{effcolor!10}6.153
      & \cellcolor{perfcolor!10}89.6 & \cellcolor{perfcolor!10}0.079 & \cellcolor{perfcolor!10}1.31 & \cellcolor{perfcolor!10}353
      & \cellcolor{effcolor!10}17.86 & \cellcolor{effcolor!10}1.46 & \cellcolor{effcolor!10}0.261
      & \cellcolor{perfcolor!10}61.3 & \cellcolor{perfcolor!10}0.502 & \cellcolor{perfcolor!10}8.32 & \cellcolor{perfcolor!10}283
      & \cellcolor{effcolor!10}157.34 & \cellcolor{effcolor!10}4.56 & \cellcolor{effcolor!10}23.548 \\
    
    PowerFast (top-swap)
      & \cellcolor{perfcolor!10}69.8 & \cellcolor{perfcolor!10}0.531 & \cellcolor{perfcolor!10}8.32 & \cellcolor{perfcolor!10}903
      & \cellcolor{effcolor!10}58.74 & \cellcolor{effcolor!10}11.92 & \cellcolor{effcolor!10}7.002
      & \cellcolor{perfcolor!10}86.8 & \cellcolor{perfcolor!10}0.077 & \cellcolor{perfcolor!10}0.92 & \cellcolor{perfcolor!10}177
      & \cellcolor{effcolor!10}9.22 & \cellcolor{effcolor!10}1.06 & \cellcolor{effcolor!10}0.098
      & \cellcolor{perfcolor!10}53.8 & \cellcolor{perfcolor!10}0.430 & \cellcolor{perfcolor!10}8.32 & \cellcolor{perfcolor!10}245
      & \cellcolor{effcolor!10}167.00 & \cellcolor{effcolor!10}5.20 & \cellcolor{effcolor!10}20.384 \\
    
    Fixed (no adapt)
      & \cellcolor{perfcolor!10}71.1 & \cellcolor{perfcolor!10}0.585 & \cellcolor{perfcolor!10}14.8 & \cellcolor{perfcolor!10}133
      & \cellcolor{effcolor!10}8.44 & \cellcolor{effcolor!10}20.82 & \cellcolor{effcolor!10}\textbf{1.757}
      & \cellcolor{perfcolor!10}79.6 & \cellcolor{perfcolor!10}0.087 & \cellcolor{perfcolor!10}14.8 & \cellcolor{perfcolor!10}111
      & \cellcolor{effcolor!10}6.28 & \cellcolor{effcolor!10}18.59 & \cellcolor{effcolor!10}1.167
      & \cellcolor{perfcolor!10}55.4 & \cellcolor{perfcolor!10}0.777 & \cellcolor{perfcolor!10}14.79 & \cellcolor{perfcolor!10}35.0
      & \cellcolor{effcolor!10}15.49 & \cellcolor{effcolor!10}14.79 & \cellcolor{effcolor!10}\textbf{5.178} \\
    
    EOT variant
      & \cellcolor{perfcolor!10}79.3 & \cellcolor{perfcolor!10}0.416 & \cellcolor{perfcolor!10}8.32 & \cellcolor{perfcolor!10}1582
      & \cellcolor{effcolor!10}64.33 & \cellcolor{effcolor!10}10.49 & \cellcolor{effcolor!10}6.748
      & \cellcolor{perfcolor!10}87.6 & \cellcolor{perfcolor!10}0.070 & \cellcolor{perfcolor!10}0.92 & \cellcolor{perfcolor!10}249
      & \cellcolor{effcolor!10}9.13 & \cellcolor{effcolor!10}1.05 & \cellcolor{effcolor!10}0.096
      & \cellcolor{perfcolor!10}60.2 & \cellcolor{perfcolor!10}0.492 & \cellcolor{perfcolor!10}8.32 & \cellcolor{perfcolor!10}288
      & \cellcolor{effcolor!10}151.77 & \cellcolor{effcolor!10}4.72 & \cellcolor{effcolor!10}23.964 \\
    
    Stealth variant
      & \cellcolor{perfcolor!10}78.4 & \cellcolor{perfcolor!10}0.425 & \cellcolor{perfcolor!10}8.32 & \cellcolor{perfcolor!10}1035
      & \cellcolor{effcolor!10}59.95 & \cellcolor{effcolor!10}10.62 & \cellcolor{effcolor!10}6.367
      & \cellcolor{perfcolor!10}89.0 & \cellcolor{perfcolor!10}0.075 & \cellcolor{perfcolor!10}0.92 & \cellcolor{perfcolor!10}188
      & \cellcolor{effcolor!10}9.55 & \cellcolor{effcolor!10}1.03 & \cellcolor{effcolor!10}0.098
      & \cellcolor{perfcolor!10}60.2 & \cellcolor{perfcolor!10}0.584 & \cellcolor{perfcolor!10}8.32 & \cellcolor{perfcolor!10}297
      & \cellcolor{effcolor!10}161.04 & \cellcolor{effcolor!10}5.07 & \cellcolor{effcolor!10}24.713 \\
    \bottomrule
  \end{tabular}%
  }
\end{table*}
}

\noindent\textbf{Effect on headline metrics.} Table~\ref{tab:combined} summarizes how each representative \emph{configuration family} performs on digital-domain attack metrics: strict success rate, target-score drop, patch area, and number of detector queries. Clear trade-offs emerge across YOLO, Faster R-CNN, and YOLOS. For instance, \texttt{gp5} matches the baseline's success rate while roughly halving the median queries on YOLO (1079$\to$529) due to shortened growth patience, with only a small area increase on Faster R-CNN. \texttt{nocx\_strict} slightly outperforms the baseline on YOLO despite removing contextual placement, while it is nearly tied on Faster R-CNN. On YOLOS, strict success is lower than on the CNN detectors but remains substantial, with \method{} reaching 59.1\%. Conversely, \texttt{fixed160} succeeds using minimal queries but requires significantly larger patches. Robustness and stealth-oriented settings (\texttt{eot10}, \texttt{stealth-strong}) maintain high success with query counts that are broadly comparable to the baseline.

\noindent\textbf{Efficiency trade-offs.}
Given consistent mAP degradation, we now assess the effort per successful attack. Table~\ref{tab:combined} combines median detector queries and patch area into the joint cost \(J_{\text{frac}} = q_{\text{med}} \times (a_{\text{med}}/100)\). On YOLO, the non-adaptive \texttt{fixed160} attains the lowest \(J\) by accepting conspicuously large patches; among \emph{adaptive} variants, \texttt{gp5} substantially reduces \(J\) relative to the baseline by halving queries without enlarging patches. On Faster R-CNN, the baseline achieves the lowest \(J\) thanks to very small patches; \texttt{powerfast} and \texttt{eot10} remain competitive. On YOLOS, \texttt{gp5} offers the best adaptive efficiency, while \texttt{fixed160} achieves the lowest overall \(J\) at the cost of a larger footprint. Overall, minimizing queries alone is insufficient if patch size remains large, highlighting the value of adaptive growth when visibility matters.

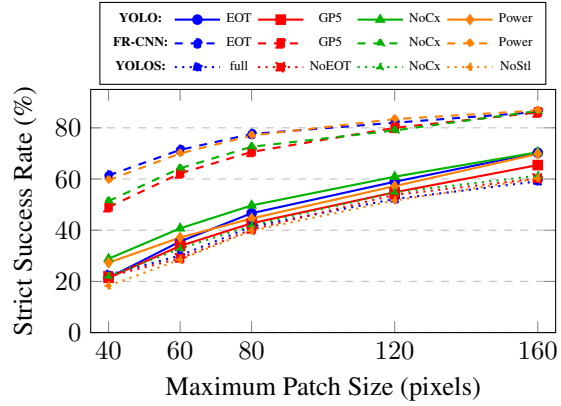
\begin{figure}[t]
\centering
\hspace*{-10mm}
\begin{tikzpicture}
\begin{axis}[
    width=0.9\columnwidth,
    height=4.8cm,
    xlabel={Maximum Patch Size (pixels)},
    ylabel={Strict Success Rate (\%)},
    xmin=35, xmax=165,
    ymin=0, ymax=95,
    xtick={40,60,80,120,160},
    legend style={font=\tiny, at={(0.98,1.36)}},
    legend columns=5,
    ymajorgrids=true,
    grid style=dashed,
    mark options={scale=0.8}, 
]
\addplot[blue, thick, mark=*, solid, forget plot] coordinates {(40,21.41) (60,35.59) (80,46.64) (120,58.95) (160,70.27)};
\addplot[red, thick, mark=square*, solid, forget plot] coordinates {(40,21.41) (60,33.86) (80,42.73) (120,54.82) (160,65.45)};
\addplot[green!70!black, thick, mark=triangle*, solid, forget plot] coordinates {(40,28.86) (60,40.73) (80,49.68) (120,60.91) (160,70.45)};
\addplot[orange, thick, mark=diamond*, solid, forget plot] coordinates {(40,27.32) (60,37.14) (80,44.59) (120,57.23) (160,69.82)};
\addplot[blue, thick, mark=*, dashed, forget plot] coordinates {(40,61.4) (60,71.4) (80,77.6) (120,82.0) (160,86.2)};
\addplot[red, thick, mark=square*, dashed, forget plot] coordinates {(40,49.0) (60,62.2) (80,70.6) (120,80.0) (160,86.0)};
\addplot[green!70!black, thick, mark=triangle*, dashed, forget plot] coordinates {(40,51.4) (60,64.2) (80,72.6) (120,79.0) (160,86.6)};
\addplot[orange, thick, mark=diamond*, dashed, forget plot] coordinates {(40,59.6) (60,70.0) (80,77.0) (120,83.4) (160,86.8)};
\addplot[blue, thick, mark=*, dotted, forget plot] coordinates {(40,22.6) (60,30.1) (80,41.4) (120,52.2) (160,59.1)};
\addplot[red, thick, mark=square*, dotted, forget plot] coordinates {(40,22.0) (60,29.0) (80,40.3) (120,53.8) (160,60.2)};
\addplot[green!70!black, thick, mark=triangle*, dotted, forget plot] coordinates {(40,22.0) (60,32.8) (80,41.9) (120,54.8) (160,61.3)};
\addplot[orange, thick, mark=diamond*, dotted, forget plot] coordinates {(40,18.3) (60,28.5) (80,39.8) (120,51.6) (160,60.2)};
\addlegendimage{empty legend}
\addlegendentry{\textbf{YOLO:}}
\addlegendimage{blue, thick, mark=*, solid}
\addlegendentry{EOT}
\addlegendimage{red, thick, mark=square*, solid}
\addlegendentry{GP5}
\addlegendimage{green!70!black, thick, mark=triangle*, solid}
\addlegendentry{NoCx}
\addlegendimage{orange, thick, mark=diamond*, solid}
\addlegendentry{Power}
\addlegendimage{empty legend}
\addlegendentry{\textbf{FR-CNN:}}
\addlegendimage{blue, thick, mark=*, dashed}
\addlegendentry{EOT}
\addlegendimage{red, thick, mark=square*, dashed}
\addlegendentry{GP5}
\addlegendimage{green!70!black, thick, mark=triangle*, dashed}
\addlegendentry{NoCx}
\addlegendimage{orange, thick, mark=diamond*, dashed}
\addlegendentry{Power}
\addlegendimage{empty legend}
\addlegendentry{\textbf{YOLOS:}}
\addlegendimage{blue, thick, mark=*, dotted}
\addlegendentry{full}
\addlegendimage{red, thick, mark=square*, dotted}
\addlegendentry{NoEOT}
\addlegendimage{green!70!black, thick, mark=triangle*, dotted}
\addlegendentry{NoCx}
\addlegendimage{orange, thick, mark=diamond*, dotted}
\addlegendentry{NoStl}
\end{axis}
\end{tikzpicture}
\caption{Cumulative success vs.\ patch size across three detector architectures. \textbf{Solid}: YOLO (one-stage CNN). \textbf{Dashed}: Faster R-CNN (two-stage CNN). \textbf{Dotted}: YOLOS (ViT-backbone transformer). Adaptive variants reach roughly 43--50\% by 80\,px and 65--70\% at 160\,px on YOLO, 70--80\% by 80\,px and 86--87\% at 160\,px on Faster R-CNN, and 40--42\% by 80\,px before plateauing near 60\% at 160\,px on YOLOS.}
\label{fig:cumsucc}
\end{figure}

\noindent\textbf{Growth dynamics.} 
Figure~\ref{fig:cumsucc} shows how success accumulates as the patch grows. On YOLO (solid), variants reach 20--30\% at 40\,px, rise to 45--50\% by 80\,px, and reach 65--70\% at full size. Faster R-CNN (dashed) climbs faster: 50--60\% at 40\,px, $>\!70\%$ by 60\,px, and plateaus near 85--87\% by 120\,px. YOLOS (dotted) follows a slower but steady trajectory, reaching about 40--42\% by 80\,px and plateauing near 60\% at 160\,px. These trends highlight the benefit of adaptive growth: many images succeed at small or moderate patch sizes, with larger patches used only when necessary. GP5's steeper early rise reflects reduced growth patience; fixed-size baselines (omitted) capture only terminal outcomes.

\subsection{Ablation Studies}
\label{sec:ablation}


\begin{table}[t]
  \centering
  \scriptsize
  \caption{Fixed-size patch ablation (digital, contextual placement).
           Efficiency cost \(J_{\text{frac}}=q_{\text{med}}\times(a_{\text{med}}/100)\).}
  \label{tab:abl_fixed}
  \resizebox{\columnwidth}{!}{%
    \begin{tabular}{@{}llccccc@{}}
      \toprule
      \textbf{Detector} & \textbf{Config} & \textbf{Strict (\%)} & \textbf{Drop$_t$} & \textbf{Area (\%)} & \textbf{Queries} & \(\mathbf{J_{\text{frac}}}\) \\
      \midrule
      \multirow{3}{*}{YOLO}
        & method full & 77.5 & 0.452 & 8.32  & 49.0 & 4.080 \\
        & fixed--160  & 71.1 & 0.585 & 14.79 & 6.0  & 0.887 \\
        & fixed--80   & 48.2 & 0.810 & 3.70  & 20.0 & \textbf{0.740} \\
      \midrule
      \multirow{3}{*}{Faster R-CNN}
        & method full & 89.7 & 0.075 & 0.92  & 8.0  & \textbf{0.074} \\
        & fixed--160  & 79.6 & 0.087 & 14.79 & 5.0  & 0.740 \\
        & fixed--80   & 66.2 & 0.102 & 3.70  & 5.0  & 0.185 \\
      \midrule
      \multirow{3}{*}{YOLOS}
        & method full & 59.1 & 0.467 & 8.32  & 290.0 & 24.089 \\
        & fixed--160  & 55.4 & 0.777 & 14.79 & 35.0  & \textbf{5.178} \\
        & fixed--80   & 36.0 & 0.974 & 3.70  & 200.0 & 7.396 \\
      \bottomrule
    \end{tabular}%
  }
\end{table}



\noindent\textbf{Fixed--size patches.}
To isolate the value of adaptive growth, we compare \method{} to fixed 160\,px and 80\,px patches. Table~\ref{tab:abl_fixed} shows that adaptive growth generally preserves higher strict success with a smaller footprint than fixed--160, while fixed-size settings reduce \(J_{\text{frac}}\) only by lowering success substantially (fixed--80) or requiring larger patches (fixed--160).

\noindent\textbf{Ablation: growth patience and query schedule.}
Reducing growth patience (\texttt{gp5}) roughly halves YOLO median queries (1079$\to$529) with essentially unchanged strict success (77.5$\to$77.7) and unchanged area (8.32\%). By contrast, \texttt{powerfast} yields a smaller query reduction (1079$\to$903) but lowers strict success (77.5$\to$69.8), indicating that shortcut schedules can trade effectiveness for efficiency.

\subsection{Physical Pilot Evaluation}
\label{sec:physical}

\begin{figure}[t]
  \centering
  \captionsetup[subfigure]{justification=centering,singlelinecheck=false}

  \begin{subfigure}[t]{0.24\columnwidth}
    \centering
    \includegraphics[width=\linewidth]{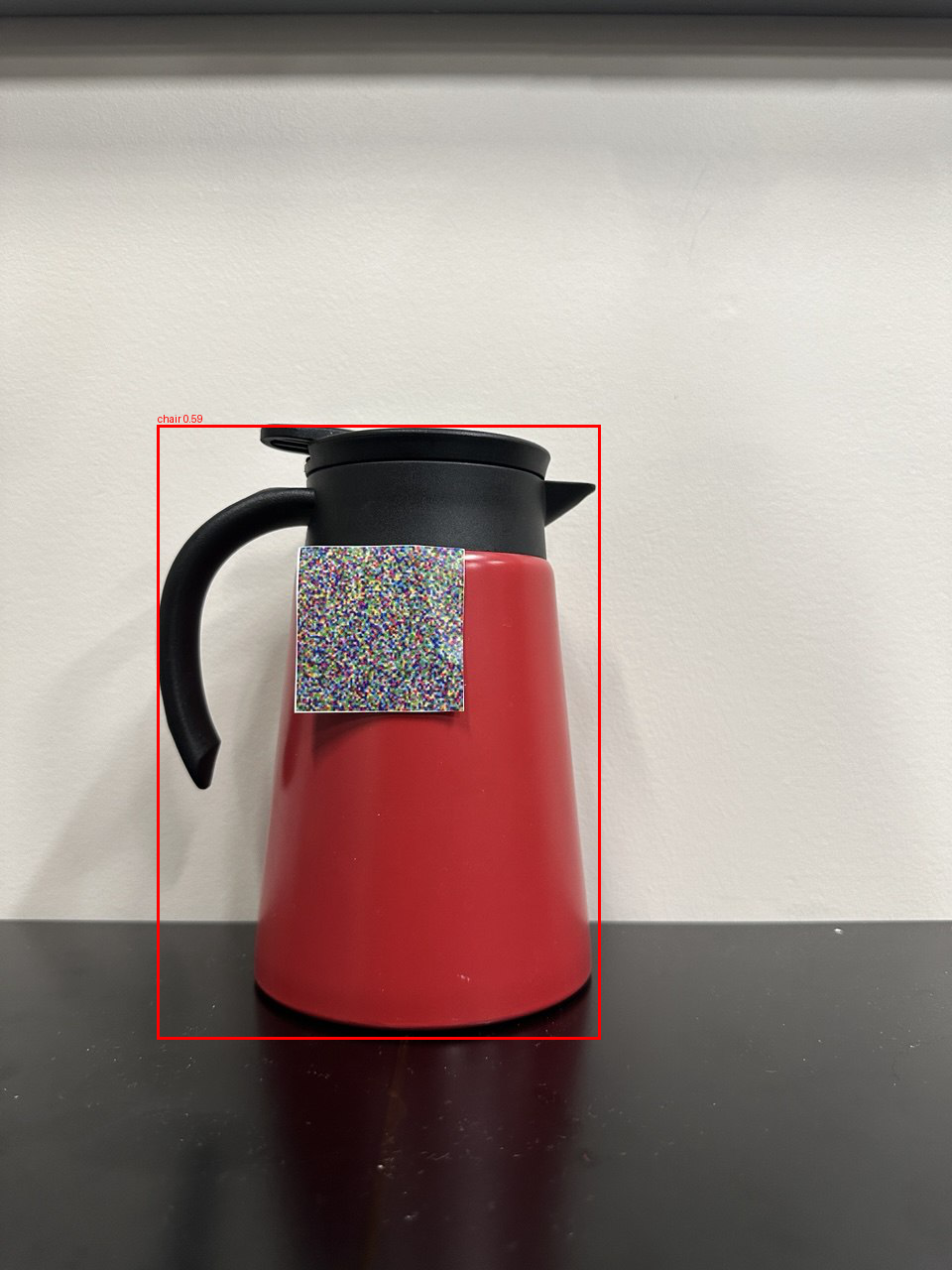}
    \caption{Frontal}
  \end{subfigure}\hfill
  \begin{subfigure}[t]{0.24\columnwidth}
    \centering
    \includegraphics[width=\linewidth]{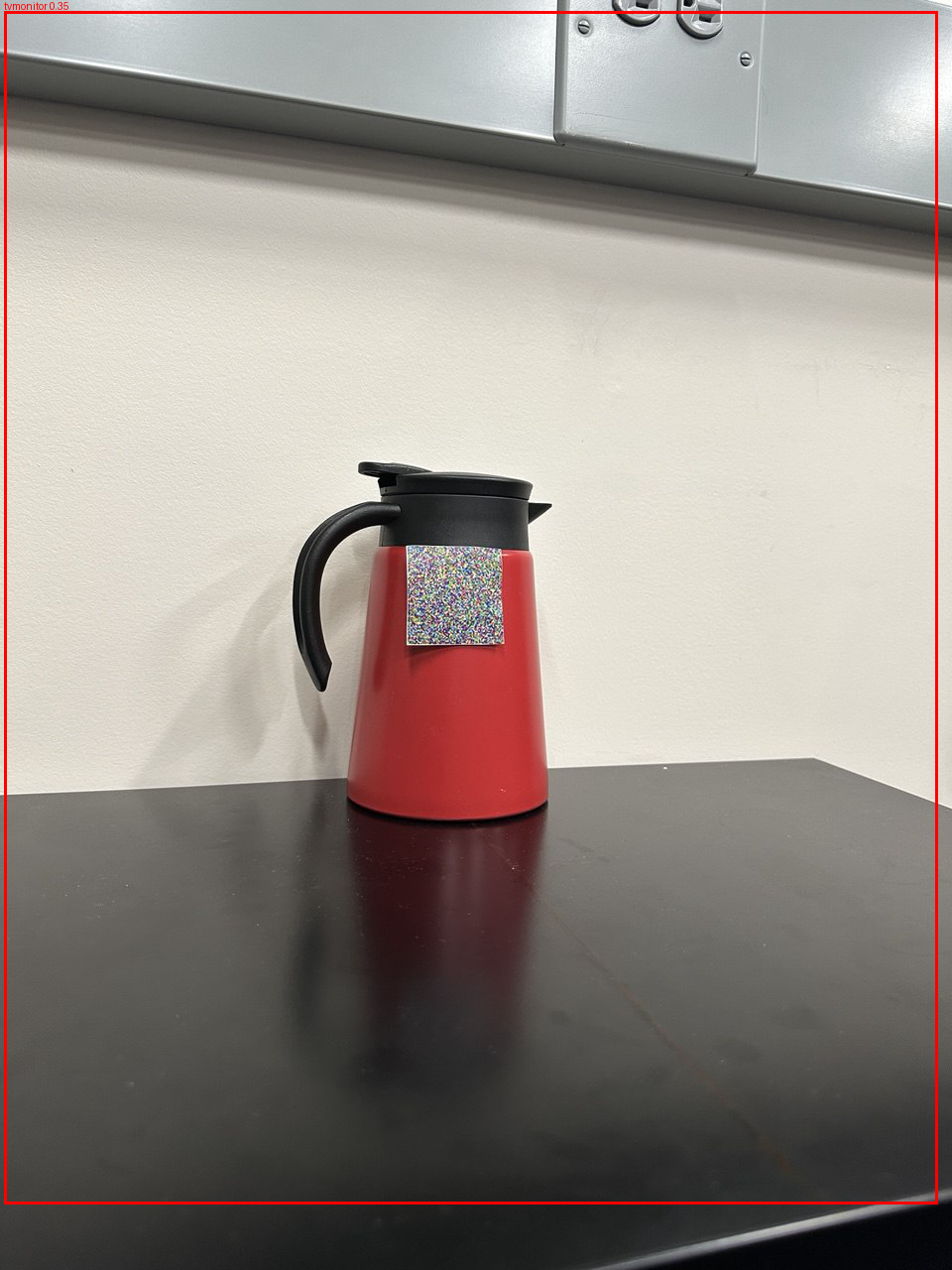}
    \caption{Tilted}
  \end{subfigure}\hfill
  \begin{subfigure}[t]{0.24\columnwidth}
    \centering
    \includegraphics[width=\linewidth]{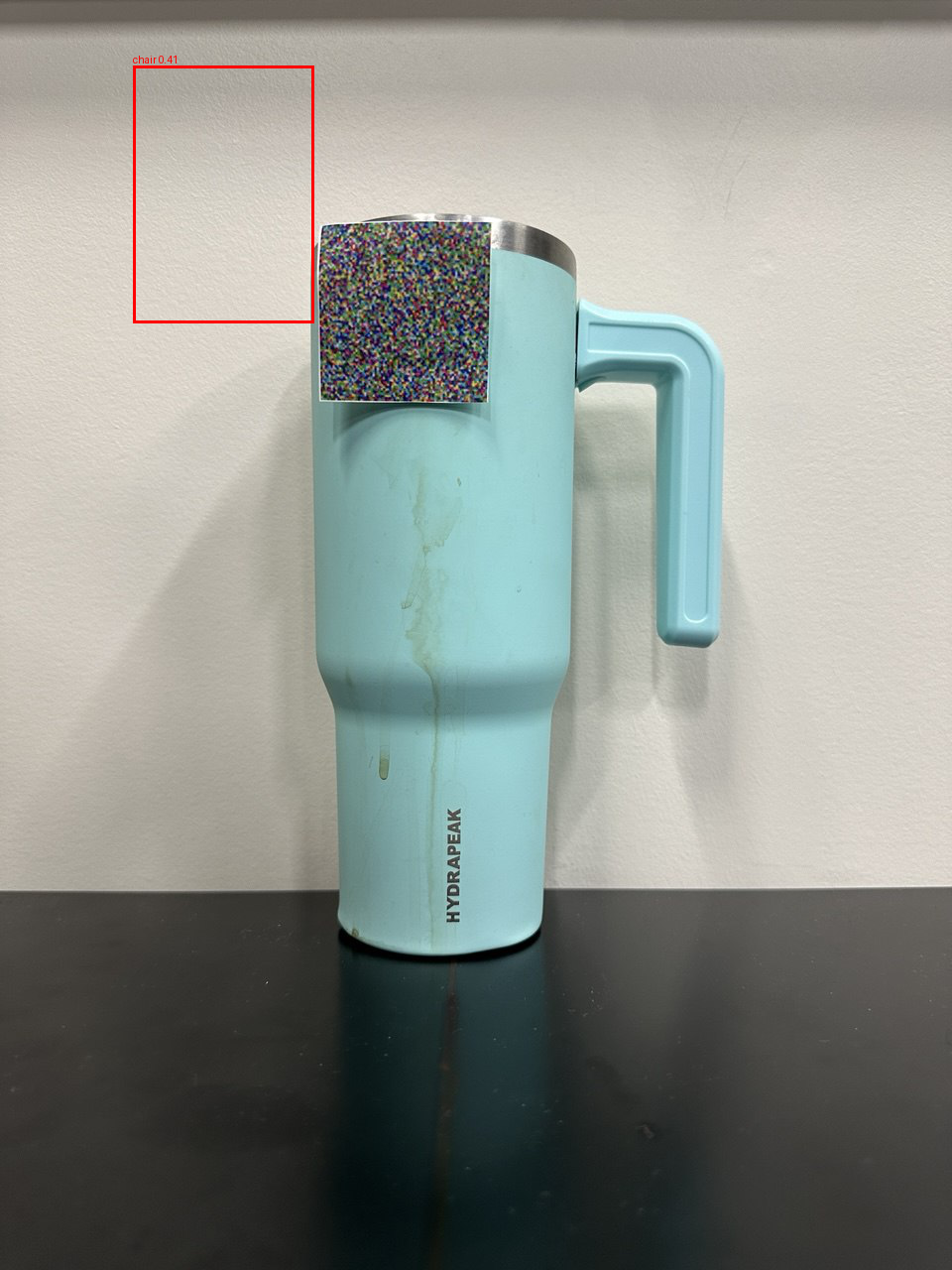}
    \caption{Transfer \#1}
  \end{subfigure}\hfill
  \begin{subfigure}[t]{0.24\columnwidth}
    \centering
    \includegraphics[width=\linewidth]{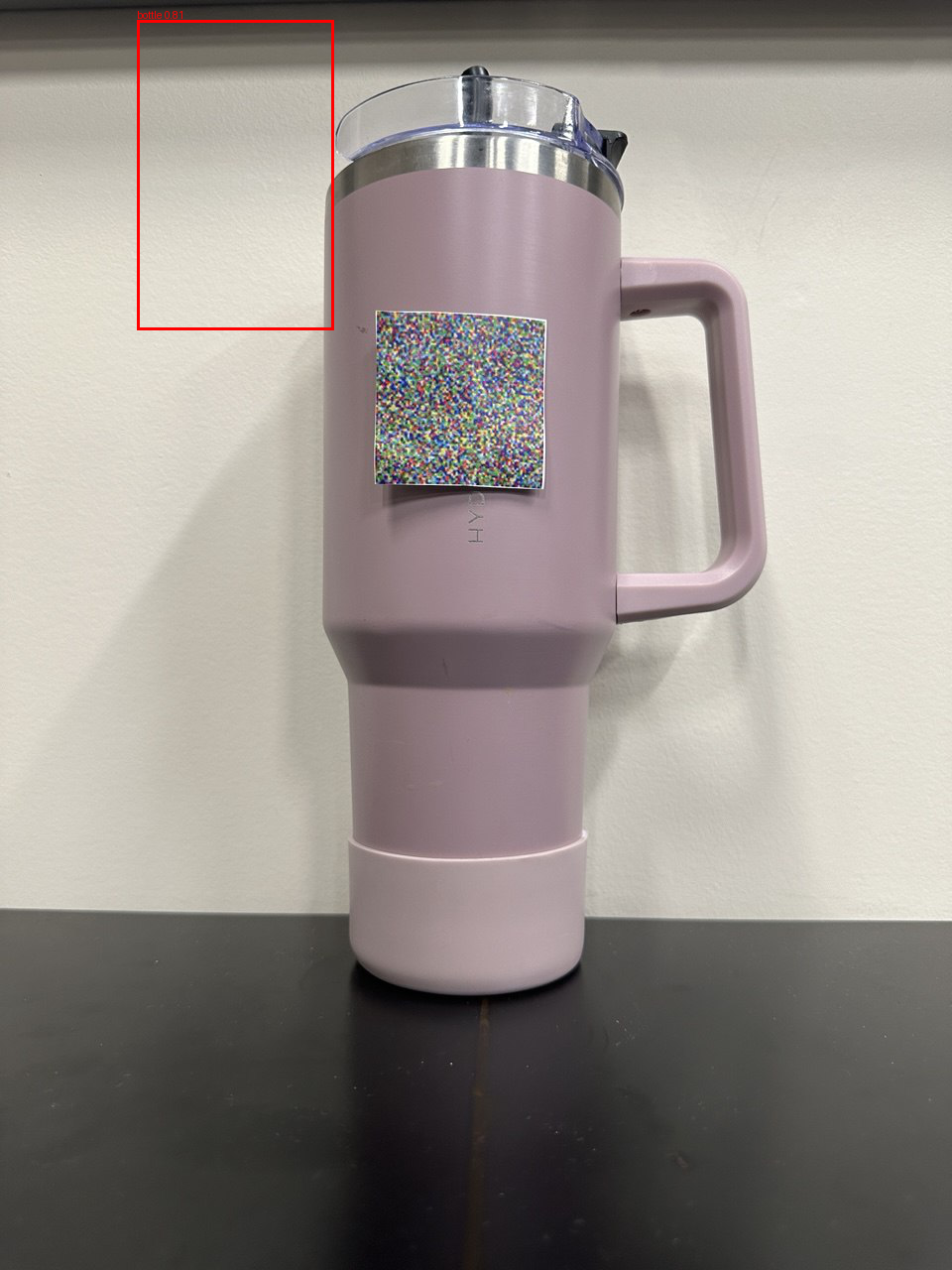}
    \caption{Transfer \#2}
  \end{subfigure}

  \caption{Representative print--capture results for the physical \method{} bottle patch. Quantitative outcomes are reported in Table~\ref{tab:phys_pilot}.}
  \label{fig:phys_bottle}
\end{figure}

To assess preliminary physical transfer, we printed a \texttt{bottle} patch optimized entirely in simulation and evaluated it under controlled changes in viewpoint, distance, and object instance. Table~\ref{tab:phys_pilot} summarizes the print--capture outcomes, and Figure~\ref{fig:phys_bottle} shows representative captures. In clean shots, YOLOv5 detected \texttt{bottle} with confidence $>\!0.90$; with the patch, the tested conditions produced misclassification, low-confidence detection, or incorrect localization.

\begin{table}[t]
  \centering
  \footnotesize
\caption{Print--capture pilot outcomes for the physical \method{} bottle patch. Patched outcomes report the top YOLOv5 prediction.}
  \label{tab:phys_pilot}
  \resizebox{\columnwidth}{!}{%
  \begin{tabular}{@{}lccccl@{}}
    \toprule
    \textbf{Condition} & \textbf{Distance} & \textbf{Azimuth} & \textbf{Object} & \textbf{Prediction} & \textbf{Outcome} \\
    \midrule
    Frontal view & 40\,cm & $0^\circ$ & Original & \texttt{chair} (0.59) & Misclassified \\
    Tilted view & 50\,cm & $10^\circ$ & Original & \texttt{tvmonitor} (0.35) & Misclassified \\
    Transfer \#1 & 54\,cm & $0^\circ$ & New \#1 & \textit{bottle} (0.41) & Wrong box / low conf. \\
    Transfer \#2 & 54\,cm & $0^\circ$ & New \#2 & \textit{bottle} (0.81) & Wrong box \\
    \bottomrule
  \end{tabular}%
  }
\end{table}

\section{Conclusion}
\label{sec:conclusion}

\method{} combines contextual Thompson-sampling placement, NES zeroth-order updates, and progress-triggered growth into a score-only attack loop. Using a strict plain-view suppression criterion and reporting EOT separately, it achieves strong suppression across CNN-based detectors and substantial suppression on a transformer-based detector, while exposing query--footprint trade-offs against fixed-size and heuristic baselines. Compact, budget-aware patches remain a realistic black-box threat, with placement playing a key role in attack efficiency.

\section*{Acknowledgments}
This work was supported in part by a grant from The BMW Group.

\bibliographystyle{IEEEbib}
\bibliography{strings,ref}

@article{ren2015faster,
  title={Faster r-cnn: Towards real-time object detection with region proposal networks},
  author={Ren, Shaoqing and He, Kaiming and Girshick, Ross and Sun, Jian},
  journal={Advances in neural information processing systems},
  volume={28},
  year={2015}
}

@article{liang2022parallel,
  title={Parallel rectangle flip attack: A query-based black-box attack against object detection},
  author={Liang, Siyuan and Wu, Baoyuan and Fan, Yanbo and Wei, Xingxing and Cao, Xiaochun},
  journal={arXiv preprint arXiv:2201.08970},
  year={2022}
}

@article{fang2021you,
  title={You only look at one sequence: Rethinking transformer in vision through object detection},
  author={Fang, Yuxin and Liao, Bencheng and Wang, Xinggang and Fang, Jiemin and Qi, Jiyang and Wu, Rui and Niu, Jianwei and Liu, Wenyu},
  journal={Advances in Neural Information Processing Systems},
  volume={34},
  pages={26183--26197},
  year={2021}
}

@inproceedings{redmon2016you,
  title={You only look once: Unified, real-time object detection},
  author={Redmon, Joseph and Divvala, Santosh and Girshick, Ross and Farhadi, Ali},
  booktitle={Proceedings of the IEEE conference on computer vision and pattern recognition},
  pages={779--788},
  year={2016}
}

@article{szegedy2013intriguing,
  title={Intriguing properties of neural networks},
  author={Szegedy, Christian and Zaremba, Wojciech and Sutskever, Ilya and Bruna, Joan and Erhan, Dumitru and Goodfellow, Ian and Fergus, Rob},
  journal={arXiv preprint arXiv:1312.6199},
  year={2013}
}

@inproceedings{yang2020patchattack,
  title={Patchattack: A black-box texture-based attack with reinforcement learning},
  author={Yang, Chenglin and Kortylewski, Adam and Xie, Cihang and Cao, Yinzhi and Yuille, Alan},
  booktitle={European Conference on Computer Vision},
  year={2020},
  organization={Springer}
}

@inproceedings{ilyas2018black,
  title={Black-box adversarial attacks with limited queries and information},
  author={Ilyas, Andrew and Engstrom, Logan and Athalye, Anish and Lin, Jessy},
  booktitle={International conference on machine learning},
  pages={2137--2146},
  year={2018},
  organization={PMLR}
}

@inproceedings{athalye2018synthesizing,
  title={Synthesizing robust adversarial examples},
  author={Athalye, Anish and Engstrom, Logan and Ilyas, Andrew and Kwok, Kevin},
  booktitle={International conference on machine learning},
  pages={284--293},
  year={2018},
  organization={PMLR}
}

@inproceedings{ilyas2018blackbox,
  title={Black-box Adversarial Attacks with Limited Queries and Information},
  author={Ilyas, Andrew and Engstrom, Logan and Athalye, Anish and Lin, Jessy},
  booktitle={ICML},
  year={2018},
  eprint={1804.08598},
  archivePrefix={arXiv}
}

@article{brown2017adversarial,
  title={Adversarial patch},
  author={Brown, Tom B and Man{\'e}, Dandelion and Roy, Aurko and Abadi, Mart{\'\i}n and Gilmer, Justin},
  journal={arXiv preprint arXiv:1712.09665},
  year={2017}
}

@article{liu2018dpatch,
  title   = {DPatch: An Adversarial Patch Attack on Object Detectors},
  author  = {Liu, Xin and Yang, Huanrui and Liu, Ziwei and Song, Linghao and Li, Hai and Chen, Yiran},
  journal = {preprint arXiv:1806.02299},
  year    = {2018}
}

@inproceedings{thys2019fooling,
  title={Fooling Automated Surveillance Cameras: Adversarial Patches to Attack Person Detection},
  author={Thys, Lien and Van Ranst, Wiebe and Goedem{\'e}, Toon},
  booktitle={CVPR Workshops},
  year={2019}
}

@inproceedings{zolfi2021translucent,
  title={The translucent patch: A physical and universal attack on object detectors},
  author={Zolfi, Alon and Kravchik, Moshe and Elovici, Yuval and Shabtai, Asaf},
  booktitle={Proceedings of the IEEE/CVF conference on computer vision and pattern recognition},
  pages={15232--15241},
  year={2021}
}

@article{lapid2023patchinvisibility,
  title={Patch of invisibility: Naturalistic physical black-box adversarial attacks on object detectors},
  author={Lapid, Raz and Mizrahi, Eylon and Sipper, Moshe},
  journal={arXiv preprint arXiv:2303.04238},
  year={2023}
}

@inproceedings{chen2017zoo,
  title={Zoo: Zeroth order optimization based black-box attacks to deep neural networks without training substitute models},
  author={Chen, Pin-Yu and Zhang, Huan and Sharma, Yash and Yi, Jinfeng and Hsieh, Cho-Jui},
  booktitle={Proceedings of the 10th ACM workshop on artificial intelligence and security},
  pages={15--26},
  year={2017}
}

@article{ilyas2019bandit,
  title={Prior convictions: Black-box adversarial attacks with bandits and priors},
  author={Ilyas, Andrew and Engstrom, Logan and Madry, Aleksander},
  journal={preprint arXiv:1807.07978},
  year={2018}
}

@inproceedings{andriushchenko2020square,
  title={Square attack: a query-efficient black-box adversarial attack via random search},
  author={Andriushchenko, Maksym and Croce, Francesco and Flammarion, Nicolas and Hein, Matthias},
  booktitle={European conference on computer vision},
  pages={484--501},
  year={2020},
  organization={Springer}
}

@article{wei2022simopt,
  title={Simultaneously optimizing perturbations and positions for black-box adversarial patch attacks},
  author={Wei, Xingxing and Guo, Ying and Yu, Jie and Zhang, Bo},
  journal={IEEE transactions on pattern analysis and machine intelligence},
  volume={45},
  number={7},
  pages={9041--9054},
  year={2022},
  publisher={IEEE}
}

@inproceedings{huang2020universal,
  title={Universal physical camouflage attacks on object detectors},
  author={Huang, Lifeng and Gao, Chengying and Zhou, Yuyin and Xie, Cihang and Yuille, Alan L and Zou, Changqing and Liu, Ning},
  booktitle={Proceedings of the IEEE/CVF conference on computer vision and pattern recognition},
  pages={720--729},
  year={2020}
}

@article{kumar2024bbpatch,
  title={BB-Patch: BlackBox Adversarial Patch-Attack using Zeroth-Order Optimization},
  author={Kumar, Satyadwyoom and Gupta, Saurabh and Buduru, Arun Balaji},
  journal={arXiv preprint arXiv:2405.06049},
  year={2024}
}

@inproceedings{eykholt2018robustphysical,
  title={Robust physical-world attacks on deep learning visual classification},
  author={Eykholt, Kevin and Evtimov, Ivan and Fernandes, Earlence and Li, Bo and Rahmati, Amir and Xiao, Chaowei and Prakash, Atul and Kohno, Tadayoshi and Song, Dawn},
  booktitle={Proceedings of the IEEE conference on computer vision and pattern recognition},
  pages={1625--1634},
  year={2018}
}

\end{document}